\documentclass{article}

\usepackage{PRIMEarxiv}

\usepackage[utf8]{inputenc} 
\usepackage[T1]{fontenc}    
\usepackage{hyperref}       
\usepackage{url}            
\usepackage{booktabs}       
\usepackage{amsfonts}       
\usepackage{nicefrac}       
\usepackage{microtype}      
\usepackage{lipsum}
\usepackage{fancyhdr}       
\usepackage{graphicx}       
\graphicspath{{media/}}     
\usepackage{wrapfig}
\usepackage{rotating}
\usepackage{pdflscape}
\usepackage{afterpage}
\usepackage{multirow}
\usepackage[utf8]{inputenc}
\usepackage{array}
\usepackage{makecell}
\usepackage{xurl}

\title{Synthetic Data in Healthcare
}

\author
{Daniel McDuff,$^{1\ast}$ Theodore Curran,$^{1}$ Achuta Kadambi$^{2}$\\
\\
\normalsize{$^{1}$University of Washington, Seattle, USA}\\
\normalsize{$^{2}$University of California, Los Angeles (UCLA), Los Angeles, USA}\\
\\
\normalsize{$^\ast$To whom correspondence should be addressed; E-mail:  dmcduff@uw.edu.}
}

\begin{document}
\maketitle

\begin{abstract}
Synthetic data are becoming a critical tool for building artificially intelligent systems. Simulators provide a way of generating data systematically and at scale. These data can then be used either exclusively, or in conjunction with real data, for training and testing systems. Synthetic data are particularly attractive in cases where the availability of ``real'' training examples might be a bottleneck. While the volume of data in healthcare is growing exponentially, creating datasets for novel tasks and/or that reflect a diverse set of conditions and causal relationships is not trivial. Furthermore, these data are highly sensitive and often patient specific. Recent research has begun to illustrate the potential for synthetic data in many areas of medicine, but no systematic review of the literature exists. In this paper, we present the cases for physical and statistical simulations for creating data and the proposed applications in healthcare and medicine. We discuss that while synthetics \emph{can} promote privacy, equity, safety and continual and causal learning, they also run the risk of introducing flaws, blind spots and propagating or exaggerating biases. 
\end{abstract}


\section{Introduction}

The volume of data generated in healthcare is growing exponentially\footnote{\url{https://www.statista.com/statistics/1037970/global-healthcare-data-volume/}} with exobytes of patient records, imaging, lab tests, physiological measurements and a plethora of other artifacts created every year. Data-driven learning methods will likely be one of the most significant tools in scientific and clinical research for many years to come~\cite{hinton2018deep,ghassemi2020review}. Artificial intelligence (AI) algorithms are gaining adoption to automate data analysis and to help improve decision-making efficiency and efficacy. In healthcare, AI has been explored in a variety of applications including detection of pathology in medical imaging (e.g., radiography, computed tomography [CT], and magnetic resonance imaging [MRI])~\cite{wang2020covid},  diagnosis of cardiovascular disease (e.g. electrocardiogram [ECG])~\cite{attia2019artificial}, analysis and prediction of health outcomes using electronic health records (EHR)~\cite{choi2016doctor,liang2019evaluation}, and information mining from medical literature~\cite{gu2021domain}.

New neural models (e.g., convolutional neural networks, long-short term models and transformers) have contributed to significant improvements across many of these tasks. These large-parameter models are designed to learn complex non-linear relationships ``end-to-end'' from data. As such, they are a product of the observations used to train them and it is widely accepted that data is perhaps the most important consideration when building a model. This is exemplified by a new field of \emph{data-centric} AI which focuses on the practice of iterating and collaborating on data used to build AI systems~\cite{mazumder2022dataperf}, with the emphasis that curating a dataset is the most effective way to improve the model.  

Obtaining, organizing, and labeling these data are expensive, laborious, and technically challenging components in the model development pipeline. Ensuring that these data have the appropriate diversity and represent the true relationships and distribution observed in practice is non-trivial. In healthcare, the collection, management, and sharing of data is complicated by privacy concerns, technical constraints, cost, and disincentives. Patient privacy and proper consent to data usage are key ethical and legal considerations under the European General Data Protection Regulations (GDPR) and the United States Health Insurance Portability and Accountability Act (HIPAA). As regulations trend towards the protection and privacy of patients, it is becoming increasingly difficult to collect sufficient volumes of labeled examples that exploit the scalability of neural models.

What if data could be ``generated'' rather than ``collected?'' Synthetic data generation, in which samples are generated by a computer model on-demand, is becoming a very powerful tool in AI model development. 
Synthetic data generation can take the form of physical models (e.g. computer simulators, computer graphics pipelines), statistical models (e.g. generative networks), or a hybrid of the two (fusion models)~\cite{de2021next}.
These models are more powerful the more precisely a practitioner can control attributes of the examples they create. Synthetic data for training AI models is a multi-billion dollar market and it is estimated that as soon as 2030~\cite{arora2022synthetic} a majority of data used to train models will be synthetic\footnote{\url{https://blogs.gartner.com/andrew_white/2021/07/24/by-2024-60-of-the-data-used-for-the-development-of-ai-and-analytics-projects-will-be-synthetically-generated/}}.
The use of simulations to augment existing datasets has been extensively explored in many areas of machine learning, including computer vision~\cite{shotton2011real,veeravasarapu2015model,haralick1992performance} and natural language processing~\cite{marzoev2020unnatural,wang2018synthetic} with promising results. Generative models are now increasingly being tested within specific applications in medicine and healthcare~\cite{hernandez2022synthetic,hahn2022contribution}. In these applications, synthetic data can be used to augment real datasets to provide supplementary examples or as the sole data for training systems. In support of the case for synthetic data in healthcare, AI models trained with synthetic data can be comparable to models trained with real data~\cite{shin2018medical}; taken a step further, AI models trained with a combination of synthetic data and real data have been shown to outperform models trained with real data alone~\cite{golany2020simgans,mcduff2020advancing}. Companies are already beginning to exploit synthetic data for commercial healthcare applications\cite{benaim2020analyzing,foraker2020spot}; however, there remain many challenges and risks in doing so.

\section{Synthetic Data Generation}

\begin{figure}
\centering
  \includegraphics[width=\textwidth]{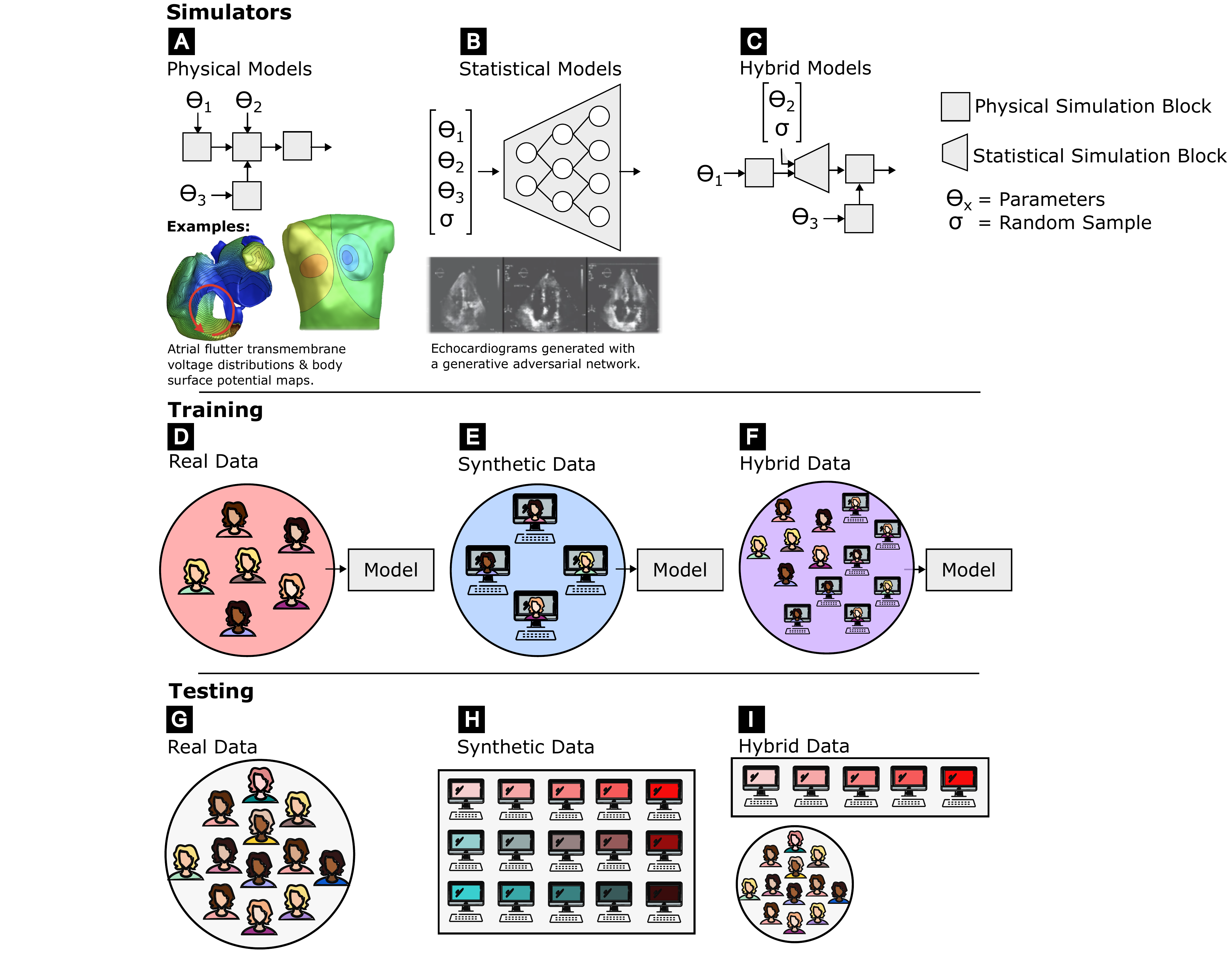}
  \caption{Simulations can be A) deterministic physical simulations, B) statistical generative simulations, or C) a hybrid of two in which physical and statistical simulations are combined.  Machine learned models can be trained using D) real, E) synthetic and F) a mixture of real and synthetic data from these simulations. Synthetic data can be used for testing/performance evaluation, to contrast G) testing on real data, H) systematically generated synthetic data and I) a combination of the two. Images of atrial flutter voltage distributions adapted from Dossel et al.~\cite{dossel2021computer} and echocardiograms adapted from Madani et al.~\cite{madani2018deep}. }
  \label{fig:syntheticdatasummary}
\end{figure}

\begin{figure}
\centering
  \includegraphics[width=0.7\textwidth]{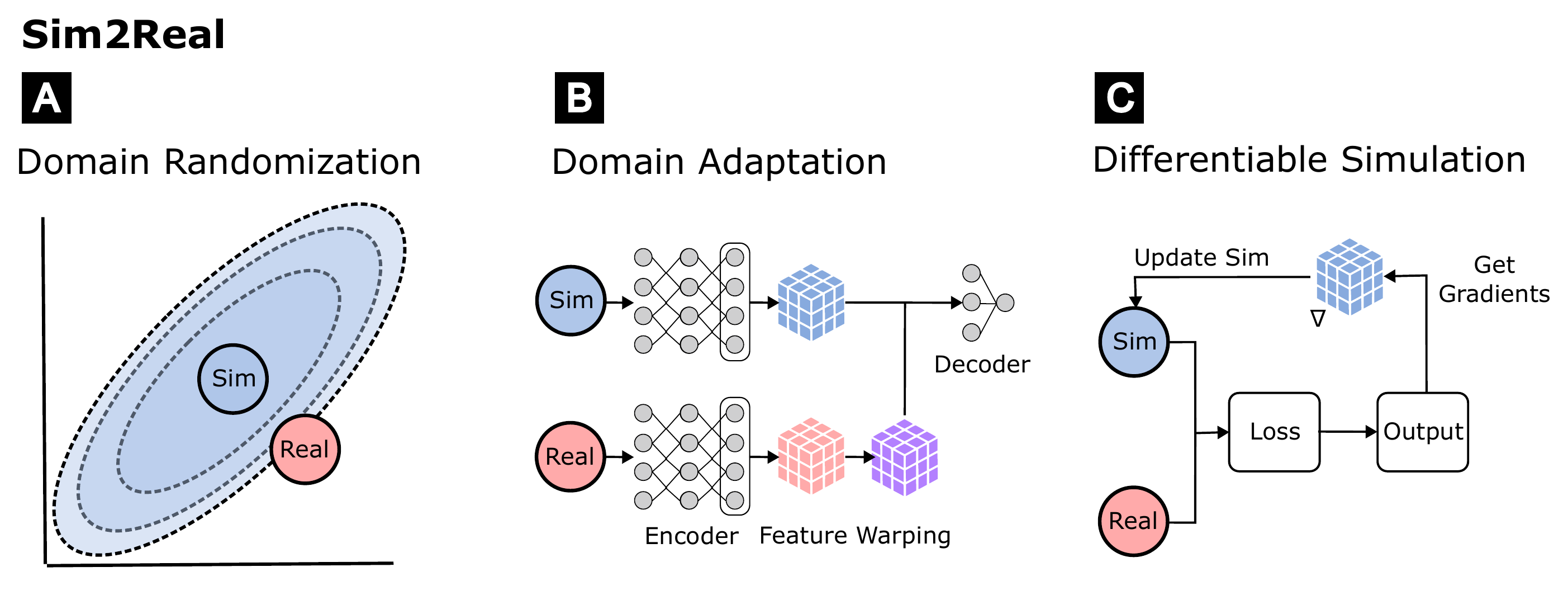}
  \caption{Methods of achieving sim2real include transfer learning approaches, such as (A) domain randomization or B) domain adaptation. Because it is sometimes difficult to transfer models from simulations to real data, another strategy is to improve the realism of simulators in a hybrid physics/data manner, through methods like C) differentiable simulation.}
  \label{fig:sim2real}
\end{figure}

\subsection{Physical Models}

Physical models (see Fig. 1A) require knowledge of how to parameterize the data generation process. Computer graphics engines simulate the physics of the real world and are a way of creating renderings of physical environments, structures, and events. For example, to design a model for simulating the cardiac blood volume pulse would require parameterization of the fluid mechanics of the cardiovascular system. Detailed computer simulations with large numbers of parameters are often expensive and time consuming to create and can require significant computational resources. Given the need to manually parameterize the model, these simulations need to be designed for specific tasks and may not necessarily generalize, meaning that researchers in different domains need to design independent pipelines.
Physical models have the advantage of being interpretable and having known constraints on the samples they can and cannot produce; in this regard they present a potentially safer option for creating data. However, \emph{model mismatch} can occur where the real world is often highly complex and a physical model does not reflect all the variance that might be observed. A simple model might be acceptable if there are known unknowns which can be mitigated using guardrails around the use of a model; however, unknown unknowns present a much more significant risk to the practical utility of these models.

\subsection{Statistical Models}

Statistical generative models capture a probabilistic representation of a dataset from which samples can be drawn. These are often referred to as ``generative AI'' (ChatGPT is an example of a statistical generative model\footnote{https://openai.com/blog/chatgpt}) and are trained to mimic the distribution of a dataset. Generative adversarial networks (GANs)~\cite{goodfellow2020generative}, variational autoencoders (VAEs)~\cite{kingma2013auto}, flow models~\cite{kingma2018glow}, and diffusion models~\cite{sohl2015deep,ho2020denoising,dhariwal2021diffusion} are all examples of generative architectures. For a review of deep generative models for synthetic data see Eigenschink et al.~\cite{eigenschink2021deep}. As these are trained models, albeit generally unsupervised, they are a function of the data used to create them. Compared to physical models, statistical models more often faithfully capture the distribution of data observed; however, this is only a strength if the training data is representative. High resolution images with very realistic contents can be simulated using these models~\cite{karras2019style}. 
If data is available, statistical models are substantially less complex to create than some physical simulators.  For example, rendering highly realistic videos of the human body requires elaborate physically-grounded graphics models, whereas a generative model can be trained to produce realistic videos with much less effort. Statistical models are generally much less ``controllable'' or ``understandable'' than physical models because each parameter, or group of parameters (e.g., layer in a network), does not interact in a logical manner based on physics or any other theoretical bases. For example, for the cardiac blood volume pulse, a generative model might produce waveforms that have very reasonable physiologic properties, but it might be hard to identify which parameter(s) map to the velocity of the forward wave or the heart rate variability. Supervised learning can be used to create models that are controllable (\emph{conditioned}) along certain dimensions; however, more work is still required to ensure the reliability of samples from such models in applications such as medicine.

\subsection{Sim2Real}

Ideally, models trained on synthetic data should generalize to real-world conditions. A sim2real \emph{domain gap} exists when models perform differently in the real domain as compared to the synthetic domain. In order to minimize this gap, one can build more realistic simulators, but simulating our physical world is hard to do. Three frameworks to minimize domain gap are illustrated in Figure~\ref{fig:sim2real}. \emph{Domain randomization} perturbs the generation of simulated data, thereby shifting the distribution of the generated data. As shown in Figure~\ref{fig:sim2real}a, after a sufficiently large distribution shift, the real world will fall under the range of data that is simulated. For example, a model that produces RGB images from one model of digital camera might fail to produce images that match another model of camera; however, shifting the pixel distribution may address this discrepancy. Another technique known as \emph{domain adaptation} is illustrated in Figure~\ref{fig:sim2real}b. A model trained in the simulated world will ordinarily perform better using features learned from simulated data. Domain adaptation seeks to warp features from real world data to match features from simulators. Both domain adaptation and domain randomization are a form of transfer learning, which involves training a model on data from one domain that will be tested on data from other domain. A third technique known as \emph{differentiable simulation} takes a more direct approach in trying to make the simulator more realistic. Differentiable simulation aims to form a simulator whose parameters can be optimized. One can then use learning to set the optimal parameters of the simulator. Imagine a robot learning to pick up shiny objects with complex reflective properties. An initial simulator might not be able to accurately model shiny appearance, but after collecting a few images and seeing how the simulator and real data differ, one can learn how to optimize the simulator. For example, the simulator could be extended to model reflections more effectively. In healthcare, it is critical that any form of synthetic patients generalize to real data, and Sim2Real techniques will likely have a role to play. Examples include domain adaptation to improve the performance of MRI tumor segmentation~\cite{sahu2020endo} and coherence tomography (CT) material decomposition models trained on synthetic data~\cite{abascal2021material}. 

\section{The Case for Synthetic Data in Healthcare}

Complete, accessible, and representative patient data is essential to clinical and scientific research, internal operational evaluation and quality improvement, and potential AI-driven tools such as clinical diagnostics and decision support. Such data is elusive; the following is a discussion of the primary supporting arguments for the use of synthetic data in healthcare. 

\textbf{Data Sharing and Privacy.} The barriers to data sharing in healthcare are numerous and at times enigmatic. Patient data is restricted primarily for privacy concerns enforced by regulations such as HIPAA and GDPR. Even within the same healthcare organization, sharing of patient data is cumbersome due to policies that understandably prioritize privacy. In addition, healthcare systems historically have lacked incentives to share data, given the expense involved to acquire data, such that data has been treated in a proprietary manner. Recent state-level regulatory efforts have created financial incentives to increase data sharing between hospitals and clinics in the hope of reducing costs and improving patient care, but the scope appears to be limited to patient-care purposes and the ultimate impact is not yet known\cite{juhn2022incentivizing}. Similarly, the medical research community lacks a culture of data sharing, due in part to misaligned professional incentives\cite{bierer2017data}, data formats being optimized for clinical care and not analysis, and in large part to patient privacy concerns and the subsequent cost, time, and effort involved to share data ethically and legally. There are a few rare exceptions of databases of clinical patient data made available to the research community, but even these are not enough due to limitations in population demographics, diseases, and types of data (e.g., MIMIC-IV~\cite{johnson2023mimic}, eICU~\cite{pollard2018eicu}, AmsterdamUMCdb~\cite{thoral2021sharing}, HiRID~\cite{faltys2021hirid}, PIC~\cite{zeng2020pic}).

Although synthetic data cannot address all barriers to sharing data, it can overcome perhaps the largest barrier, patient privacy. The current standard of de-identified patient data has been shown to be vulnerable to malicious methods, such as matching to other known records to expose the original identifiers\cite{janmey2018re,yoo2018risks}. Synthetic data provides a unique approach to identify protection because samples are generated not from an individual, but rather a blend of characteristics across individuals~\cite{rankin2020reliability}.  In other industries, privacy-preserving datasets make it easier for data to be shared and used by commercial and academic entities, encouraging collaboration, extending impact and enabling secondary analyses~\cite{azizi2021can}. A recently published general (i.e. ``task-agnostic'') AI method for preserving data privacy may have implications for healthcare data ~\cite{singh2022decouple}. Computer graphics models and other physical simulations can be constructed without direct use of real data. Statistical models with differentiable privacy guarantees can be created~\cite{jordon2018pate}. Even at the individual level, synthetic data can be created without exposing genuine samples, thereby creating a "digital twin." Ultimately, whether synthetic data, created from models trained with real data, falls under protected health information will be left to regulators and institutional review boards. While not specific to health data, a couple of general tools for synthesizing ML data that are considered to be privacy preserving can be found in Liu et al.~\cite{liu2021iterative} and Zhang et al.~\cite{zhang2021privsyn}.

\textbf{Fairness and Equity.} In the AI field, ``fairness,'' though not concretely defined~\cite{verma2018fairness}, generally refers to a model or distribution of data that accurately reflects the cohort in question without bias in the context of legal and social norms; similarly, in the healthcare field, “equity” generally refers to an even distribution of demographics, namely age, gender, ethnicity, and race, in regards to human subjects research or patient care. It is critical to ensure that a medical intervention does not disadvantage segments of the human population. For medical devices in particular, ensuring equity is a complex problem. As Kadambi~\cite{kadambi2021achieving} explains, the fairness of medical devices is not just an AI problem but must be studied at the physical layer (how a sensor senses data); the AI layer (how an algorithm processes data); and an interpretation layer (how a human would interpret the outputs). While synthetics are not a "quick fix," they can help mitigate biases and improve fairness and equity across multiple layers. For example, consider recent evidence that pulse oximeters do not perform as well on darker skin tones~\cite{okunlola2022pulse}. An AI model can be more equitable if the dataset includes optically challenging subjects. Unfortunately, it can be hard to collect a high volume of real subjects with extreme skin tones. Here, synthetics can be generated with diverse phenotypes. An AI model trained on a diverse set of synthetics has been shown to offer improvements in equity~\cite{mcduff2022using,wang2022synthetic}. Somewhat counter-intuitively, an equitable algorithm is often a better algorithm for every stakeholder. Not only does inclusion of minority phenotypes improve model performance for minorities and the population, but also for majority groups, a phenomenon referred to by Chari et al. as ``minority inclusion for majority enhancement'' (MIME)~\cite{chari2022mime}. In the synthetic world it might be possible to create synthetic humans that blend multiple difficult phenotypes, thereby allowing an AI to learn better models. Such a dataset can be difficult or impossible to obtain in the real world, especially for rare diseases. 

\textbf{Rare Events.} In healthcare, there is a plurality of rare events such as diseases, syndromes, extreme values, or any other outlier from a normal distribution. So much so that machine learning methods are often tailored to detect those samples~\cite{li2010learning,fernando2021deep}. However, collecting samples of these events for training algorithms is challenging, frequently meaning that datasets have substantial class imbalance~\cite{li2010learning}. Simulators can be used to create synthetic examples of rare events and thereby reduce the lack of sensitivity of a model trained with class imbalance. Examples include detecting arrhythmia from PPG or ECG~\cite{kiranyaz2017personalized} or cancers in medical images~\cite{chen2021synthetic}. Synthetic data of rare events can be used to supplement existing real data to improve AI algorithms. However, when synthesizing rare events in this way it is even more critical that simulations are reliable and do not feature confounding artifacts, otherwise a model trained on these data may simply learn to recognize synthetic samples.

\textbf{Incomplete or Inadequate Data.} Beyond rare cases, given the enormous amount of data required to train AI algorithms, most cases could benefit from the inclusion of synthetic data with real data, known as data augmentation~\cite{che2017boosting}. Many research papers present methods that simply use a generative AI model to create supplementary samples that are essentially permutations of real training data~\cite{shin2018medical,frid2018gan,khan2021brain}.

\textbf{Safety and Testing.} There is a significant body of literature using synthetic data for performance testing of machine learning systems~\cite{haralick1992performance,mcduff2019characterizing}. Synthetic data allow for systematic variation of parameters with the advantage that there is only the need to train and characterize the performance of the generator once and one can then evaluate many classifiers efficiently and systematically, with potentially many more variations than were used to train the generator – essentially, sampling continuously from the manifold and not just the discrete points in the training dataset.
However, if the simulator is biased, this can propagate systemic inequalities that exist in the real-world~\cite{caliskan2017semantics} and lead to increased risks for already disadvantaged groups. A model may perform poorly on populations that are minorities within the training set and test performance may not be truely representative.

There are certain interventions, experiments and data collections that are unethical and/or unsafe for human subjects research.  Synthetic data enables evaluation of methods in safety-critical contexts without risk to patients or other research subjects. For instance, a sepsis-alert system can be tested on synthetic data for fine-tuning prior to deployment. While these approaches are suitable for understanding some limitations of a model in the healthcare domain, there are many questions as to how appropriate performance testing with synthetic data could be. These questions involve the performance of synthetic data on the task, the trustworthiness of synthetic data and even societal norms to providing and accepting medical care based on models trained with synthetic data.

\textbf{Continual Learning.} As simulators provide ways to generate data on-demand, this means they can be used for continual learning.  If a weakness or bias is discovered in a model, more samples from that domain or class could be created. If the environment or conditions under which a model is to be deployed change then a model can be updated. Significantly, synthetic data can often be generated very rapidly thus enabling a model to be updated quickly, minimizing the impact of changes in conditions on the performance. Comparatively, if real-world clinical data is required this can be a long process with institutional review boards, patient recruitment, data collection and model re-training. Once again, the concept of digital twins, means that data resembling a specific individual could be created and a personalized model fine-tuned before the model is deployed for that patient.

\textbf{Explainable AI.} Adoption of AI in healthcare will depend largely on clinicians, who are effectively end users for clinical applications. Clinicians make decisions based on a mental model trained by an established theoretical basis of pathophysiology, mechanisms of action for medications, practice experience, and knowledge of current literature and clinical trials. In contrast, AI often draws inferences which are not fully understood or explained by conventional means. Although the general process of AI is understood, the exact parameters and associations that carry weight are often times unknown and are virtually unknowable if many parameters are included. The issue of explainable AI in healthcare is complicated, as Ghassemi et al shed doubt that current methods exist to satisfy clinicians and regulators~\cite{ghassemi2021false}. Synthetic data could be used to test our understanding of a model, by altering or omitting parameters presumed to be important, and seeing if the output has changed. Synthetic data from physical models can be used to help explain a model by generating data that is interpretable in a manner familiar to clinicians, such as viewable tabular data, images, or ECGs. It is human nature to trust what is understood, and mistrust what is misunderstood. Thus, for an AI to not only be validated, but also explainable in terms familiar to clinicians is essential to successful adoption.

\textbf{Causal Models.} The explainability of AI includes examination of the cause-and-effect relationships in a model, for which there should be
a plausible mechanism based on established theory. Causal and counterfactual reasoning and inference could make a significant impact in safety critical scenarios in medicine and healthcare, where data-driven prediction models can mistakenly be used to draw causal inferences~\cite{prosperi2020causal} where understanding more than correlations are an important component. Researchers have used video and synthetic datasets for causal reasoning~\cite{fu1991causim,mcduff2022causalcity}, because simulations, specifically physical simulations, have a generating mechanism that is known. Knowing the causal structure of underlying data is a huge advantage when creating training data for training and testing causal models.

\begin{figure}
\centering
  \includegraphics[width=0.9\textwidth]{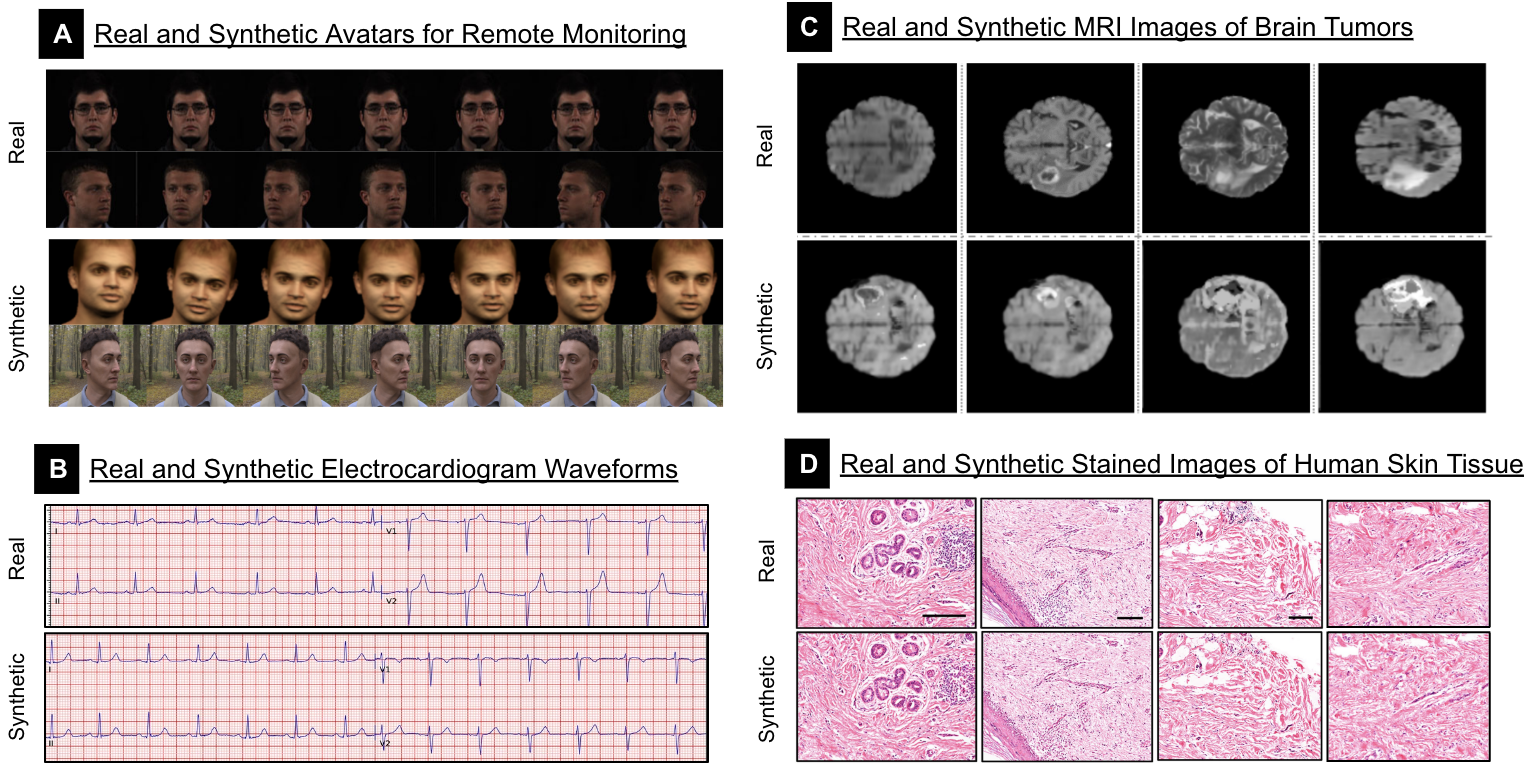}
  \caption{A) Real and synthetic patient videos for remote monitoring~\cite{estepp2014recovering,wang2022synthetic,mcduff2022scamps}. B) Real and synthetic electrocardiogram~\cite{thambawita2021deepfake}  C) Real and synthesized MRI images of brain tumors~\cite{shin2018medical}. D) Real and synthetic hematoxylin and eosin stains from TA-PARS nonradiative absorption contrast images~\cite{boktor2022virtual}.
}
  \label{fig:syntheticexamples}
\end{figure}



\section{Applications of Synthetic Data in Healthcare}

In modern healthcare systems, patient data is stored in the electronic health record (EHR). The EHR contains a complex array of data, from unstructured data, including imaging and natural language records, to structured data, including continuous, categorical, and ordinal variables, all of which exhibit linear and nonlinear interactions. Most data is a time series, in which the sequence and time interval between data points is crucial to the causal relationship and effect size. Data is stored in the form of tables, images, or natural language records, the design of which has historically prioritized clinical care and billing over data analysis.
Owing to the aforementioned advantages, synthetic data are being used in domains as diverse as cardiology, dermatology, histology and clinical psychology (see Table~\ref{tab:summary} for examples). Statistical generative AI models are the most common tool, though they may not be suitable for some applications~\cite{chen2021synthetic}. 
We will now discuss examples of applications of synthetic data in healthcare and the joint conclusions that stem from these works.

\begin{table}
	\caption{Example simulations for creating data for medical AI systems.}
	\label{tab:methods}
	\centering
	\footnotesize
	\setlength\tabcolsep{6pt} 
	\begin{tabular}{lp{3cm}p{3cm}p{5.5cm}}
	\toprule
	  \textbf{Application} & \textbf{Modality} & \textbf{Simulator} & \textbf{Examples} \\
	\hline \hline
	  \multirow{4}{*}{Cardiology}  & \multirow{2}{*}{Cardiac Waveforms} & Dynamical Models & ECG~\cite{mcsharry2003dynamical}, PPG~\cite{martin2013stochastic}, PCG~\cite{almasi2011dynamical} \\
        & & Statistical & ECG~\cite{golany2020simgans} \\ \cline{2-4}
         & \multirow{2}{*}{Cardiac Imaging} & Physical Model & Echocardiograms \cite{alessandrini2015pipeline,prakosa2012generation} \\ 
         &  & Statistical & Echocardiograms \cite{tiago2022data,gilbert2021generating}, Cardiac MRI~\cite{amirrajab2020xcat} \\ \cline{2-4}
         \hline
        Dermatology & Skin Lesions & Statistical & Images of skin conditions~\cite{ghorbani2020dermgan}  \\ \hline
        Gastroenterology & Endoscopy & Physical Model & Monocular endoscopy depth-estimation~\cite{mahmood2018unsupervised} \\ \hline
        Neurology & Brain Imaging &  Physical \& Statistical Model & Brain tumor image segmentation~\cite{menze2014multimodal} \\ \hline
        Ophthalmology & Retinopathy & Statistical Model & Retinal images for retinopathy of prematurity~\cite{coyner2022synthetic} \\ \hline
        Infectious Diseases & Detection & Statistical Model & Chest radiographs for COVID-19 det.~\cite{wang2020covid,waheed2020covidgan} \\ \hline
        Histology & Detection & Statistical Model & Detection of renal cell carcinoma~\cite{chen2021synthetic}, Colorization of skin tissue images~\cite{boktor2022virtual} \\ \hline
        Clinical Psychology & Patient Records & Statistical Model & Classifying mental health diagnoses~\cite{ive2020generation} \\
    \bottomrule
  \end{tabular}
  \footnotesize
  \\
  ECG = Electrocardiogram, PPG = Photoplethysmogram, PCG = Phonocardiogram
  \label{tab:summary}
\end{table}

\textbf{Structured Data}. The structured, or tabular, data in an EHR includes patient demographics, past medical history, medications, diagnoses, allergies, vital signs, fluid balance, laboratory results, microbiology data, pathology data, and some data related to procedures and imaging (i.e. not the image itself). Benaim et al illustrate how synthetic structured data can be utilized for clinical observational studies. Their aim was to validate the use of a commercial AI platform, MDClone, to generate synthetic structured data from real EHR data while maintaining known associations \cite{benaim2020analyzing}. They investigated rates of proton pump inhibitor prescription for gastroprotection in high risk patients following discharge, door-to-balloon time for cases of ST-elevation myocardial infarction as a predictor of heart failure and mortality at 180 days, blood urea nitrogen levels during admission as a predictor of 3-year mortality in patients with heart failure, the risk of acute kidney injury following MRI versus CT, and the risk of hypoglycemia for inpatients treated with glargine versus detemir insulin. Subsequently, Foraker et al. investigated three scenarios to validate the same AI platform in a different healthcare system \cite{foraker2020spot}. They created synthetic data from real data underlying the pediatric risk mortality (PRISM) score in a pediatric intensive care unit, an inpatient sepsis prediction model, and a public health tool for tracking rates of chlamydia. The studies demonstrated the platform could generate synthetic data with similar characteristics to real data, and importantly, the results derived from synthetic data were statistically similar to real data results. These results relied on data sets where proportionally the number of  patients is much larger than the number of variables used, and with large sample sizes to reduce the effect of censored data~\cite{benaim2020analyzing}. These studies were observational and seemingly intended for sharing data for retrospective analysis only.

Synthetic data modeled from a data set of over 1 billion records from critically ill patients (Amsterdam UMCdb~\cite{thoral2021sharing}) was used to compare methods for synthetic data generation and patient re-identification. Their analysis highlighted the need not only for high quality synthetic data generation methods but also good, robust metrics for evaluating such data~\cite{jordon2021hide}.

\textbf{Natural language records}. An EHR contains an enormous amount of natural language text within notes written by healthcare providers – the quality and accuracy of which is subjective, but in its ideal form provides a distilled narrative that highlights the most crucial parts of a patient's course and expected path forward. This is particularly important for mental health records which largely depend on unstructured natural language text more than structured data. Synthetic natural language data as training data can yield classification results that are comparable to the original results, as Ive et al. show that a model trained with synthetic discharge summaries could accurately predict diagnoses and phenotypes of mental health patients~\cite{ive2020generation}.  

Advancements made in natural language processing in other industries is promising for healthcare, for which synthetic natural language text may be of benefit. The recent popularity of ChatGPT illustrates the profound capability, impact, and scale of a modern language model. Such language models depend on unrestricted access to an enormous amount of natural language text scoured from the internet. Given the current limitations imposed by privacy protections, it is difficult for a similar language model for clinical decision support to be developed using natural language records from an EHR. A current beta of CogStack Foresight powered by MedGPT shows promise for a probabilistic diagnostic tool from raw text, but the model training was limited to just over one million patient records mostly from a single hospital system.\footnote{https://cogstack.org/cogstack-foresight-beta-launched/} Synthetic natural language text generated from real records may be able to train or at least provide augmented training data to improve such a model for a broader patient population.

\textbf{Physiological Measurements.} One of the domains in which synthetic patient data have been adopted most widely is in processing of physiological signals and downstream classification. Physiological measurement refers to any unstructured continuous data stream, such as a 12-lead ECG and ECG telemetry, or waveforms related to hemodynamics. In particular, researchers have investigated the synthesis of cardiac signals, namely ECG's, as well as phonocardiograms and photoplethysmograms (PPG)~\cite{mcsharry2003dynamical,golany2020simgans,almasi2011dynamical}.
For instance, parametric dynamical models are effective at capturing properties of ECG's~\cite{mcsharry2003dynamical,sayadi2010synthetic}, allowing for the generation of synthetic ECG's of sinus bradycardia, sinus tachycardia, ventricular flutter, atrial fibrillation and ventricular tachycardia. 
Importantly, models trained with real ECG's augmented with synthetic ECG's have shown improved accuracy of rhythm diagnosis \cite{golany2020simgans,skandarani2021generative}. Dynamical statistical models using a system
of ordinary differential equations (ODEs) incorporated into the optimization process of a generative AI, offer the ability to control the dynamic properties of waveforms while benefiting from the data-driven nature of generative AI models~\cite{golany2020simgans}. 
With growing interest in the relationship between physiologic signals (e.g., pulse arrival time computed from ECG and PPG as a predictor of blood pressure) future work should consider synthetics that model the relationships between measurements, rather than single measurements alone.

Camera measurement of physiological signals benefits from the relative ubiquity of cameras and their ability to capture measurements without physical contact~\cite{mcduff2021camera}. However, synthesizing video is more complex than one-dimensional waveforms. Graphics-based models can be used to create a diversity of physical models~\cite{mcduff2022scamps,paruchuri2023motion}. However, these do not resemble ``photo-realistic'' data yet. Statistical models provide a more accessible approach that does not require an elaborate synthetic pipeline~\cite{wang2022synthetic}. \emph{Semi-synthetic} methods have been proposed where only portions of the input data are synthesized, meaning that one can rely on real samples for hard-to-simulate portions of the data space but supplement this with synthetic training data for samples that are easier to simulate. For example, in PPG the skin tone affects waveform accuracy due to the optical properties of skin. In~\cite{ba2022style}, real patients were augmented with style transfer to mimic different skin tones to create semi-synthetics for more equitable vital sign detection.

\textbf{Medical Imaging.} Medical imaging has begun to adopt AI methods to automate portions of the clinical assessment pipeline. Optimistic claims that these models would quickly reach human parity and even replace human experts (e.g., radiologists)~\cite{langlotz2019will} have not turned out to be correct. Nevertheless, there are many promising examples supporting the use of machine learning to improve the insights obtained from these data.
Synthetic image generation has been successfully applied to improve models classifying malignant cells in histopathology slides~\cite{chen2021synthetic}, skin lesions in photographs~\cite{ghorbani2020dermgan}, detecting COVID-19 from chest radiographs~\cite{waheed2020covidgan,wang2020covid}, classifying liver lesions from CT scans~\cite{frid2018gan} and segmenting brain tumors in MRI data~\cite{menze2014multimodal}. In the majority of work, statistical generative AI models (most commonly GANs) have been employed~\cite{waheed2020covidgan,ghorbani2020dermgan} (see Table~\ref{tab:summary}), but there are also examples of physical models and hybrid models being used in cases when the physical models are well characterized~\cite{zhao2019data,billot2021synthseg}. In particular, hybrid methods are used to amplify synthetic data through physics-informed data augmentations. For example Zhao et al.~\cite{zhao2019data} compute augmentations on synthetic data that are consistent with spatial deformation fields that arise in the context of anatomy and MRI image acquisition. Amirrajab et al.~\cite{amirrajab2020xcat} used a heart simulator combined with a generative model to generate synthetic cardiac MR images. In a leap of faith, Zhang et al.~\cite{zhang2018translating} used a generative AI model to translate 3D cardiac MR images to 2D CT format whilst theoretically sparing radiation and radiocontrast dye exposure. With new forms of imaging devices continuing to proliferate the need for efficient ways to create training datasets will only continue~\cite{nowara2022seeing}. Synthetic data will likely serve an increasingly important role. 



\section{Challenges and Risks}

While these examples highlight the proliferation of synthetic data in artificial intelligence for medicine and healthcare; they also raise concerns about the vulnerabilities of the software and the challenges of current policy~\cite{chen2021synthetic}. Synthetic data are very attractive but are also associated with real risks, many of which have been scantly addressed in the current literature and regulations owing to the novelty and expediency of adoption.

\textbf{Flaws and Limits in the Simulation Engine.} The most significant technical challenge associated with simulators is creating samples that reflect the true complexity of the real world and determining whether it is possible to understand and describe the limitations of the simulation. With physical models, these limitations are often more transparent because those models are built with a known, or partially known, deterministic structure and set of parameters. Accurate physical models that simulate a sensor (e.g., such as an endoscope or camera) and an
anatomically-realistic part of the body (e.g., colon or the face)~\cite{mahmood2018unsupervised,mcduff2022scamps} present a principled approach that could be more compliant with existing clinical regulations~\cite{chen2021synthetic}. 
However, with statistical generative AI models, limitations can be much harder to define.  Because statistical models are trained it is unclear whether they can effectively generate samples that fall outside the distribution of the training data. This means that they are inherently limited by the data used to train them and raises questions about how they complement the existing data that remain to be fully answered. For example, language models can stitch together phrases without necessarily generating any new insights~\cite{bender2021dangers}. Statistical synthetic data generation methods for structured data from an EHR are prone to generating data that violates expected relationships, for example, between sex and diseases~\cite{yan2022multifaceted}. A review of synthetic data generation of structured tabular data in palliative care concluded that despite the number of papers and novel methods, there was limited consensus on how to evaluate the performance of synthetic data generation methods~\cite{hahn2022contribution}. These problems will hinder adoption and make it difficult for synthetic data to garner trust.

\textbf{Unknown Unknowns.} When modeling a phenomena there are known knowns (e.g., a parameterizable relationship or property), known unknowns (e.g., a relationship that is known to exist but cannot be quantified for the lack of empirical data or proof) and unknown unknowns (e.g., a relationship that may or may not exist but is overlooked by the model developer). It is the last of these that are the most problematic. Providing confidence that a simulator has sufficient fidelity that unknown unknowns will be inconsequential is non-trivial and may be impossible to prove. To date, researchers have typically relied on empirical evidence that a model trained on synthetic data generalizes to real test data. However, performance of simulations on real data is only one piece of the puzzle. In the future, one may use simulations not only for performance gains, but to help distill or discover underlying relationships that were not known previously~\cite{schmidt2009distilling,chari2019visual}. Simulators may need input from an oracle, a ``human-in-the-loop'', to identify failures in these cases~\cite{lakkaraju2017identifying,chen2021synthetic}.

\textbf{Lack of Standards and Regulations for Evaluating Models Trained with Synthetic Data.} The US FDA AI-based software
as a medical device (AI-SaMD) action plan\footnote{https://www.fda.gov/medical-devices/software-medical-device-samd/artificial-intelligence-and-machine-learning-software-medical-device} does not discuss simulated, synthetic or augmented data. Chen et al. discuss the limitations of this action plan with regard to synthetic data in detail~\cite{chen2019validity}, concluding that new regulations ``need to be developed and specifically adapted for different use cases.'' 
Despite the rapidly growing interest and potential positive use cases of synthetic data, there is no clear legislation surrounding their use in healthcare~\cite{arora2022synthetic}. A problem that contributed to this issue is that new methods are often disclosed without sufficient transparency and documentation. The consequence of being that it is impossible to carry out objective evaluations~\cite{walters2020assessing}.

\textbf{Lack of Representation and Bias.} Both physical model-based and statistical methods can be at risk of not representing the true diversity of patients and conditions. 
The lack of representation often leads to biases in models as learned generative models produces statistics relative to the underlying data distribution~\cite{grover2019bias} and large-parameter models have been found to amplify bias~\cite{bender2021dangers}.  There is a body of literature to help address these biases~\cite{hardt2016equality,van2021decaf}, but tools for model creators are still limited.  Model cards can help document the data used to create a generative model so that practitioners who employ these models have an understanding of the limitations and flaws.~\cite{mitchell2019model}.

\textbf{Data Leakage.} Although synthetic data offers great potential for patient privacy protection, there is risk for synthetic generation pipelines to leak personal information~\cite{chen2020gan,jordon2021hide}.  There are no methods for strictly determining whether synthetic data is sufficiently different from real samples to be considered anonymized~\cite{arora2022synthetic}. There is a privacy-utility trade off for any simulator, in which the benefits of privacy protections must be weighed against diminishing utility for the proposed task \cite{sayadi2010synthetic}. While this is particularly true for statistical models, it is also true for parametric models that rely on elements involving identifiable human data such as facial or body images~\cite{mcduff2021camera}.  Simulators that are overfit or those trained from a small data set of real patients are especially vulnerable to reverse engineering methods that may expose identifiable information within the original data \cite{chari2022mime}. For particularly sensitive cases, individual records can be excluded from a model without damaging model fidelity, provided the sample size is large enough.  Leakage of private health information (PHI) via malicious attacks or unintentional data generation that matches a specific individual is problematic, leading Chen et al.~\cite{chen2021synthetic} to suggest that statistical generative models are not suitable for some applications of synthetic generation in healthcare. Such sentiment motivated Jordan et al. to conduct a novel challenge towards a protected synthetic dataset created by a generative model with a real ICU dataset, by pitting ``hiders,'' teams attempting to create protected synthetic datasets, against ``seekers,'' teams attempting to re-identify patients based on the synthetic data~\cite{jordon2021hide}.

\section{Conclusion}

Simulators are attractive as they enable on-demand generation of samples for training and/or testing of data-intensive machine learning methods. Synthetic data can allow for sharing of data that otherwise would violate patient privacy protections. In some cases, synthetic data may even substitute for samples that might be unethical or unsafe to collect, or simply impractical due to the rarity of many diseases.
However, there are several challenges and risks associated with the use of synthetic data in clinical applications. Flaws in the data generation process could lead to flaws in a model. In particular, limits on a simulator mean certain types of realistic samples cannot be created, biases can lead to poor representation in the resulting data, the potential for unknown unknowns means synthetic data cannot be fully trusted, and personal data leakage remains a risk as a simulator can expose sensitive data. Much more work is needed to address these challenges and we observed many examples in which synthetic data were applied rather naively.
In many ways, simulations serve to amplify current AI models: they can amplify the size and diversity of a dataset; however we must also acknowledge that they can also amplify the errors and biases of AI pipelines if not used correctly. Ensuring that models trained on simulations are tested on real data, and verified to be useful, is as close to a silver bullet as we can have to ensure that simulations are of benefit. This reinforces the need for closed-loop collaboration between the practitioners generating synthetic data and clinical experts who are collecting and drawing inference on real data. This returns us to the close links between the work in data-centric AI and simulation. More tools and standards that support assessing and improving the quality of simulations will be necessary if they are to be adopted in practical medical applications.

\section*{Acknowledgments}

The authors would like to acknowledge  Aditya Gupta (UCLA), Praneeth Vepakomma (MIT) and Celso de Melo (US Army Research Lab) for their comments on earlier version of this manuscript. Achuta Kadambi partially supported by an NSF CAREER Award (IIS-2046737), Army Young Investigator Award, and DARPA Young Faculty Award.

\bibliographystyle{unsrt}
\bibliography{references}

\end{document}